\documentclass[conference]{IEEEtran}
\IEEEoverridecommandlockouts
\usepackage{cite}
\usepackage{amsmath,amssymb,amsfonts}
\usepackage[linesnumbered,ruled]{algorithm2e}
\usepackage{graphicx}
\usepackage{textcomp}
\usepackage{xcolor}
\usepackage{tabu}
\usepackage{multirow}
\usepackage{float}
\usepackage{url}
\usepackage{amsthm}
\usepackage{subfig}
\def\BibTeX{{\rm B\kern-.05em{\sc i\kern-.025em b}\kern-.08em
    T\kern-.1667em\lower.7ex\hbox{E}\kern-.125emX}}

\begin{document}

\title{Metamorphic Testing of a Deep Learning based Forecaster}

\author{
\IEEEauthorblockN{Anurag Dwarakanath\IEEEauthorrefmark{1}, Manish Ahuja\IEEEauthorrefmark{1}, Sanjay Podder\IEEEauthorrefmark{1}, Silja Vinu\IEEEauthorrefmark{2},  Arijit Naskar\IEEEauthorrefmark{2}, Koushik MV\IEEEauthorrefmark{2} }
\IEEEauthorblockA{\IEEEauthorrefmark{1} \textit{Accenture Labs}\\
Bangalore, India \\
\{anurag.dwarakanath, manish.a.ahuja, sanjay.podder\}@accenture.com}
\IEEEauthorblockA{ \IEEEauthorrefmark{2} \textit{Accenture}\\
Bangalore, India \\
\{silja.vinu, arijit.naskar, mv.koushik\}@accenture.com}
}

\IEEEoverridecommandlockouts
\IEEEpubid{\begin{minipage}{\textwidth}\ \\ \\ This is the author final version. \\ Paper published at the 2019 IEEE/ACM 4th International Workshop on \\ Metamorphic Testing (MET). Published version:\\ https://dl.acm.org/citation.cfm?id=3340651.3340662 \\
978-1-7281-2235-9/19/\$31.00~\copyright2019 IEEE \\
DOI 10.1109/MET.2019.00014 
\end{minipage}}


\maketitle

\begin{abstract}
In this paper, we present the Metamorphic Testing of an in-use deep learning based forecasting application. The application looks at the past data of system characteristics (e.g. `memory allocation') to predict outages in the future. We focus on two statistical / machine learning based components - a) detection of co-relation between system characteristics and b) estimating the future value of a system characteristic using an LSTM (a deep learning architecture). In total, 19 Metamorphic Relations have been developed and we provide proofs \& algorithms where applicable. We evaluated our method through two settings. In the first, we executed the relations on the actual application and uncovered 8 issues not known before. Second, we generated hypothetical bugs, through Mutation Testing, on a reference implementation of the LSTM based forecaster and found that 65.9\% of the bugs were caught through the relations. 
\end{abstract}

\begin{IEEEkeywords}
Metamorphic Testing of Deep Learning, Co-relation co-efficient, LSTM
\end{IEEEkeywords}

\section{Introduction}
With the recent success of machine learning (ML) based applications (such as computer vision, speech recognition, etc), we expect most businesses, in the near future, to use some form of ML in their applications. However, a key problem with ML applications is its inherent difficulty to test\cite{Dwarakanath}. This difficulty arises from the fact that a) the input space is extremely large; and b) for most inputs, it is highly non-trivial to know the expected results.


One such ML based application, used in practice, is an `Outage Predictor' (henceforth referred to as the OP application). This application trains a model on the past characteristics of a system, such as `memory allocation', processor usage', and predicts the future values of these characteristics. An outage is predicted if the forecast for any of these characteristics is beyond a threshold. We focus on two components of the OP application - a) the generation of statistical co-relation between the characteristics; and b) forecasting the value of a system characteristic using a deep learning based LSTM (long-short-term-memory) architecture.
\newline
\newline
\newline
\newline
\newline
Testing the OP application is extremely challenging since the input space, which comprises of the range of real valued features, is extremely large. For example, if the system has a memory capacity of $128\ Gb$, the feature of `memory usage' can vary from $0$ to $137438953472$ i.e. ($1.3 * 10^{11}$). Further, there are multiple other such features making the combinations exponentially larger. Once a forecast is made by the application, checking whether the forecast is correct, cannot be done by hand. The current state of practice, to test the forecasts, is to collect a large amount of validation data and measure the deviance of the forecast from the actual values. Such testing is grossly insufficient and can miss many potential issues in an ML application. We detail the problems in the current state of practice in Section \ref{related}.   

In this work, we investigate the application of Metamorphic Testing to identify issues in the OP application. We have developed a set of 19 Metamorphic Relations (MRs) for the testing of the application and we include proofs where possible. These MRs were executed on the application and 8 previously unknown issues were uncovered. Note that these issues were found even though the application had completed the standard practice of measuring the performance on validation data. To further assess the efficacy of the MRs, we created a set of hypothetical bugs on a reference implementation of the LSTM based forecaster using the concept of Mutation Testing. Of the 44 bugs that were generated, the MRs were able to catch 29. Overall, the results show the strong practical applicability of Metamorphic Testing. 


There are several pieces of novelty in our work. To the best of our knowledge, our work is the first to develop MRs for an LSTM based deep learning architecture. Two of the MRs introduces new algorithms - one which evaluates the critical hyper-parameter of how many past instances of data should be used for training; and another which evaluates the robustness of the LSTM forecaster using adversarial techniques. We have also presented MRs for Pearson's statistical co-relation co-efficient. Further, we report results of the MRs on an in-use application as well as the overall efficacy of the MRs on a set of hypothetical bugs. 

The paper is structured as follows. We present the Metamorphic Testing of the co-relation co-efficient generation in Section \ref{corel}. The MRs for the LSTM based forecaster is presented in Section \ref{lstm}. We present the results in Section \ref{results} which includes the results from the in-use application and the efficacy on the hypothetical bugs. Related work is in Section \ref{related} and we conclude in Section \ref{conclusion}.

\section{Module 1: co-relation co-efficient generation} \label{corel}
The co-relation co-efficient generation module of the OP application is used to identify the set of system characteristics that are useful to predict the target (another system characteristic). For example, let the system characteristic that needs to be forecasted be denoted as $y$. Let the set of all system characteristics, be denoted as $x_1, x_2, x_3 ... x_n$. Module 1 orders the set of characteristics  $x_1, x_2, x_3 ... x_n$ in decreasing order of its ability to predict $y$. Using this order, the user can select which of the system characteristics (among $x_1, x_2, x_3 ... x_n$) is to be considered as the input for the LSTM module.

The co-relation co-efficient between an input feature $x$ and the target $y$ is computed using Pearson's co-relation co-efficient and is shown in Equation \ref{pearsonCor}. Here, the pair of data points are represented as $x_i$ and $y_i$. The mean of $x$ and $y$ are represented as $\mu_x$ and $\mu_y$. 

\begin{equation}
	\label{pearsonCor}
	r_{xy} = \frac{\sum_{i=1}^{n} (x_i - \mu_x)(y_i - \mu_y)}{\sqrt{\sum_{i=1}^{n}(x_i - \mu_x)^2} \sqrt{\sum_{i=1}^{n}(y_i - \mu_y)^2}}
\end{equation}


We now present the set of Metamorphic Relations (MRs) to test the implementation of Pearson's co-relation co-efficient.

\subsection{MR-1: bounds}
The Pearson's co-relation co-efficient, by definition, is a value between -1 and +1. In any instance, if the co-relation is found to be beyond these values, a bug is present. Such a relation is said to belong to the class of Identity relations (and Metamorphic relations generalise this notion) \cite{segura2016survey}.

\[
-1 \leq r \leq +1
\]

\subsection{MR-2: change location of features in input data}
A property of the co-relation co-efficient, $r$, is that it is symmetric - i.e. $r_{xy} = r_{yx}$. The symmetric property is directly due to the commutative property of multiplication (i.e. $a * b = b * a$) as can be seen from Equation \ref{pearsonCor}.

There can be multiple ways in which the location can be changed. A good practice is to change the columns spanning the boundaries (i.e. either the beginning or the end of the input data columns). 

\subsection{MR-3: change location of data-points together in both features}
Another property of the co-relation co-efficient is its invariance to the location of pairs of data-points. For example, a pair of data points $(x_1, y_1)$ can be moved to a different location $x_{500}, y_{500}$ and $r$ would not change. From Equation \ref{pearsonCor}, changing the locations of the data-points does not change the denominator as the standard deviation, $\sigma$, remains the same. Further, since $\sum$ is commutative ($a + b = b + a$), the numerator does not change as well.

Again, a number of variations of this relation can be tried where the location of the data-points can be put at numerous places. A good practice is to move the location to the boundary cases - the start (top) and end (bottom) of the input data. 

\subsection{MR-4: duplicate values of a feature into a new feature}
The maximum value of the co-relation co-efficient of $+1$ is obtained when there is a perfect co-variance between the two features. This can be achieved by introducing a new feature whose values are an exact duplicate of another feature. This can be easily proved by replacing $y$ with $x$ in Equation \ref{pearsonCor}:

\begin{proof}
\[
	\begin{aligned}
	r_{xx} & = \frac{\sum_{i=1}^{n} (x_i - \mu_x)(x_i - \mu_x)}{\sqrt{\sum_{i=1}^{n}(x_i - \mu_x)^2} \sqrt{\sum_{i=1}^{n}(x_i - \mu_x)^2}} \\
	& = \frac{\sum_{i=1}^{n} (x_i - \mu_x)^2}{{\sum_{i=1}^{n}(x_i - \mu_x)^2} } = 1\\ 
	\end{aligned}
\]
\end{proof}

\subsection{MR-5: multiply values of a feature by $-1$ and introduce as a new feature}
To obtain a perfect negative co-relation, we introduce a new feature but negate all the values. Let $x$ be duplicated to $-x$. Then we have $\mu_{-x} = -\mu_x$.

\begin{proof}
\[
	\begin{aligned}
		r_{x-x} & = \frac{\sum_{i=1}^{n} (x_i - \mu_x)(-x_i + \mu_x)}{\sqrt{\sum_{i=1}^{n}(x_i - \mu_x)^2} \sqrt{\sum_{i=1}^{n}(-x_i + \mu_x)^2}} \\
            & = \frac{-1 * \sum_{i=1}^{n} (x_i - \mu_x)(x_i - \mu_x)}{\sqrt{\sum_{i=1}^{n}(x_i - \mu_x)^2} \sqrt{ \sum_{i=1}^{n}(-1 (x_i - \mu_x))^2}} \\
						& = \frac{-1 * \sum_{i=1}^{n} (x_i - \mu_x)^2} {\sum_{i=1}^{n}(x_i - \mu_x)^2} = -1 \\
	\end{aligned}
\]
\end{proof}

\subsection{MR-6: linear scaling of values of a feature}
The co-relation co-efficient is invariant to linear scaling of the variables $x$ and $y$ - i.e. if we create a new feature $z = ay + b$ for any constants $a$ and $b$, then $r_{xy} = r_{xz}$. Because of the scaling, we have $\mu_z = \frac{1}{n} \sum_{i=1}^{i=n}ay_i + b = \frac{1}{n} a \sum_{i=1}^{i=n} y_i + bn = a\mu_y + b$. This test can be performed with multiple values of $a$ and $b$ (both positive and negative $\neq 0$).

\begin{proof}
\[
	\begin{aligned}
		r_{xz} & = \frac{\sum_{i=1}^{n} (x_i - \mu_x)(ay_i + b - a\mu_y - b)}{\sqrt{\sum_{i=1}^{n}(x_i - \mu_x)^2} \sqrt{\sum_{i=1}^{n}(ay_i + b - a\mu_y - b)^2}}\\
		& = \frac{a \sum_{i=1}^{n} (x_i - \mu_x)(y_i  - \mu_y)}{\sqrt{\sum_{i=1}^{n}(x_i - \mu_x)^2} \sqrt{a^2\sum_{i=1}^{n}(y_i - \mu_y)^2}} \\
		& = r_{xy}
	\end{aligned}
\]
\end{proof}


\subsection{MR-7: introduce new features such that the co-relation co-efficient computes as 0}
The following data, $x=\{1, 0, -1, 0\}\  y={0, 1, 0, -1}$ has non-zero standard deviation but zero co-relation co-efficient. This data can be used to ensure the application is able to handle the case of $0$ co-relation between features.  The metamorphic relation thus checks that the application works in similar fashion whether the data has $0$ co-relation co-efficient or not.

\subsection{MR-8: introduce a new feature with a variance of 0}

From Equation \ref{pearsonCor}, it can be seen that the co-relation co-efficient is undefined when either of the standard deviations is zero ($\sigma_x =0\ or\ \sigma_y = 0$). To test how the application handles this case (there should be no crash and an appropriate message should be displayed), introduce a new feature with a constant value (e.g. $1$) as shown below.

\begin{figure}[H]
\centering
\caption{Introduce a feature with a constant value so that the standard deviation is 0.}
\label{inputData_newFeature_zerostddev}
	\includegraphics[height=1.5in]{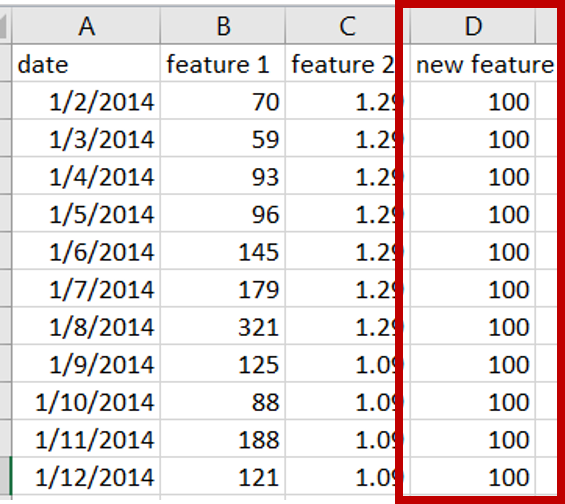}
\end{figure}

\subsection{MR-9: (Best Practice) introduce a pair of new data points that are clearly outliers.}
The co-relation coefficient is significantly affected by outliers. For example, consider the two data sets in Figure \ref{inputData_outlier}:

\begin{figure}[H]
\centering
\caption{A single large outlier has changed the results from a large co-relation ($\approx 1$) to no co-relation ($\approx 0$).}
\label{inputData_outlier}
	\includegraphics[width=2in]{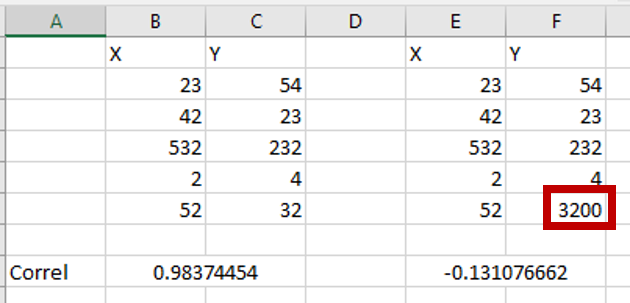}
\end{figure}

The application should remove such outliers before computing the co-relation co-efficient. To test this, introduce a new data-point which is clearly an outlier (e.g. 1000 * max value). The application should produce the co-relation co-efficient which is the same without the outlier. Appropriate warnings should also be generated.

\subsection{MR-10: introduce missing values}
This test introduces a set of missing values into the data-points. The correlation co-efficient should be computed to the same value with the introduction of a few missing values have been introduced in the data.

\section{Module 2: LSTM based forecaster} \label{lstm}
Using the output of Module-1, the user can select a set of system characteristics to be used as the predictors to estimate the forecast of another system characteristic (the target). The OP application now trains a RNN (Recurrent Neural Network) using LSTM cells (long-short-term-memory) with the values of the predictors as the input and the values of the target as the output. 


To build the set of tests for such a system, we will briefly describe the typical structure of the code which implements a RNN for forecasting using LSTM cells. We have also made a reference implementation\footnote{\url{https://github.com/anuragbms/Sales-forecasting-with-RNNs/blob/master/sales_forecasting_on_kaggle_data.ipynb}} using TensorFlow and a sales forecasting data-set\footnote{\url{https://www.kaggle.com/tevecsystems/retail-sales-forecasting}}. We will also use the reference application to generate hypothetical bugs through Mutation Testing. 

\begin{figure}[H]
\centering
	\subfloat[Mock training data depicting the sales of a particular item.]{%
		\includegraphics[width=1.5in]{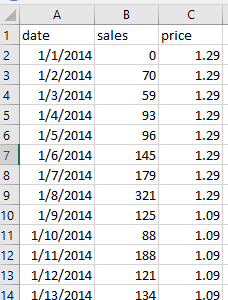}
	}%
	\hspace{18pt}%
	\subfloat[Mock validation data depicting the sales of a particular item at a different point in time.]{%
	\includegraphics[width=1.5in]{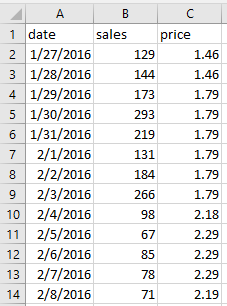}
	}%
	\caption{Sample training and validation data used in the reference implementation.}
	\label{mockTrainVaidateData}
\end{figure}

\begin{figure}[H]
\centering
	\includegraphics[width=3in]{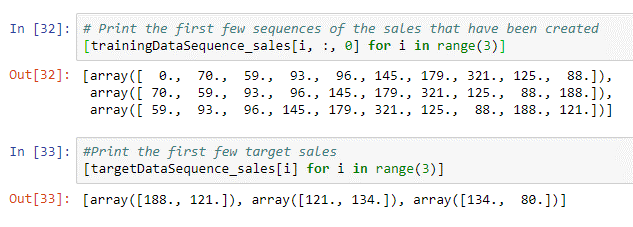}
	\caption{An example of an input data sequence and corresponding target.}
	\label{dataSequences}
\end{figure}

\paragraph{Data processing}
The training data consists of a contiguous set of values (along the time dimension) as shown in Figure \ref{mockTrainVaidateData}. However, this data needs to be transformed into a set of sequences and corresponding targets for the RNN to train. An example of the data sequences and corresponding targets is shown in Figure \ref{dataSequences}. In the figure, the values of a variable (e.g. sales of an item) is depicted for $10$ contiguous days. The target is the value of the sales for the next $2$ days. 

Going forward, we will denote the number of time steps which forms the input sequence as TIME\_STEPS and the number of time steps for which forecasts are made as NUMBER\_OF\_DAYS\_TO\_FORECAST. In the figure above, TIME\_STEPS$=10$ and NUMBER\_OF\_DAYS\_TO\_FORECAST$=2$. Such sequences are created for both the training and validation data. 

\paragraph{Outputs} The set of outputs that can be used to test the application include:
\begin{itemize}
	\item The value of the training loss after training completes
	\item The value of the validation loss when run on a trained model
	\item The value of the forecast for the first validation input
\end{itemize}

Our tests (which we will develop below) are compatible with any of the above outputs. 

\paragraph{Stochastic Nature} The training of deep learning based algorithms typically exhibit stochasticity - i.e. multiple runs on the same data can lead to slightly different results. This randomness is attributable to:

\begin{itemize}
	\item Random initialisation of the parameters of the deep learning algorithm
	\item Random shuffling of the set of training inputs
	\item Inherent randomness of the computation done on GPUs\cite{nonDeter1}\cite{nonDeter2}
\end{itemize}

In Metamorphic Testing, since we compare the outputs of multiple runs of the application, such stochastic outputs may be a problem. To counter this, we shall record the amount of variation in subsequent runs that the application exhibits without the metamorphic relations. We shall then use this default variation as a benchmark to conclude on tests executed using the relations.

Going forward, we shall denote the variation in results, without any MR as $DefaultVariation$. As an example, we have measured the $DefaultVariation$ for the reference implementation. The train file has $750$ data-points and the validation file has $187$ data-points.  In Table \ref{defaultVariation_refImpl}, the variation of the outputs for the forecast of the first validation input and the validation loss is shown.


\begin{table}[H]
	\caption{Validation forecast and validation loss for 10 runs. The difference is due to the inherent stochasticity of the algorithm.}
	\label{defaultVariation_refImpl}
	\begin{tabu}{|X[c]|X[c]|X[c]|}
		\hline
		Run \# & forecast for first validation sequence &validation loss \\
		\hline
		1 &76.453354& 0.09275999665260315 \\
		\hline
		2 &77.9922 & 0.08760423585772514 \\
		\hline
		3 &78.74777 & 0.08459033071994781 \\
		\hline
		4 &83.153175& 0.08094570226967335 \\
		\hline
		5 &81.06984& 0.08265132829546928 \\
		\hline
		6& 74.516365 & 0.0883125327527523 \\
		\hline
		7& 79.37088  & 0.08452051877975464 \\
		\hline
		8&76.03293 & 0.0892107617110014\\
		\hline
		9&33.270237 & 0.14077940210700035\\
		\hline
		10&64.52415  &0.06668609846383333 \\
		\hline
	\end{tabu}
\end{table}

The mean value, $\mu$, of the forecast computes to $72.5130901$ and the standard deviation, $\sigma$, is $14.67463148$. Assuming the number of runs is sufficient\footnote{recommended number of runs is 30}, the standard error computes to $se = \frac{\sigma}{\sqrt{number\ of\ runs}} = 4.640525931$. Thus a 95\% confidence interval would be $\mu - 1.96*se = 63.41765927$ and $\mu + 1.96*se = 81.60852093$. 

To interpret this - if in the outcome of any MR (which requires re-training of the algorithm), is outside this 95\% confidence range, we can indicate a bug in the code (with a 95\% confidence).

We now present the set of Metamorphic Relations for the computation of the forecast using an LSTM.

\subsection{MR-1: invariance of training \& validation to linear scaling of the data.}

The input training \& validation data is normalised before being passed to subsequent processing. Normalisation helps in faster convergence of stochastic gradient descent based optimisation methods. While normalisation can be done in number of different ways, the OP application uses the following:

\begin{equation}
	x_{normalised} = \frac{x - min(X)}{max(X) - min(X)}
	\label{norm}
\end{equation}

where $X$ denotes the vector (i.e. entire set of data) and $x$ denotes an individual data-point. Once, the forecasts are made, the output is re-scaled to original range using:

\begin{equation}
	x^{forecast} = (max(X) - min(X)) * x_{normalised}^{forecast} + min(X)
	\label{fcast}
\end{equation}

Test 1 specifies the following. If a constant is added to every instance of the training data \& to every instance of the validation data, the value of training loss and the value of the validation loss will approximately be the same (i.e. the only difference that would come is due to the default stochastic variation). Further, the forecast for a validation input will approximately increase by the same constant.

\begin{proof}
By adding a constant $k$ to every value of $x$ in Equation \ref{norm} we have:
\[
	\begin{aligned}
		x_{normalised} & = \frac{x + k- min(X+k)}{max(X+k) - min(X+k)} \\
		& = \frac{x + k - min(X) -k}{max(X) + k - min(X) - k} \\
		& = \frac{x - min(X)}{max(X) - min(X)}
	\end{aligned}
\]
\end{proof}

Thus, by adding a constant, there will be no change to the computation of the normalised values - i.e. the same training data as before will be used for training. Thus, except for the stochastic nature of deep learning, we expect the training loss and the validation loss to be the same.  

The addition of the constant will impact Equation \ref{fcast} as follows:

\begin{proof}
\[
	\begin{aligned}
		x^{forecast} & = (max(X+k) - min(X+k)) * x_{normalised}^{forecast}\\
		& + min(X+k) \\
		& = (max(X) - min(X)) * x_{normalised}^{forecast} \\
		&+ min(X) + k\\
	\end{aligned}
\]
\end{proof}

\begin{figure}[H]
\centering
	\subfloat[Mock training data with a constant (309) added to every data-point.]{%
		\includegraphics[width=1.5in]{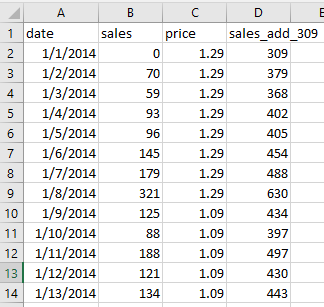}
	}%
	\hspace{18pt}%
	\subfloat[Mock validation data with a constant (309) added to every data-point.]{%
	\includegraphics[width=1.5in]{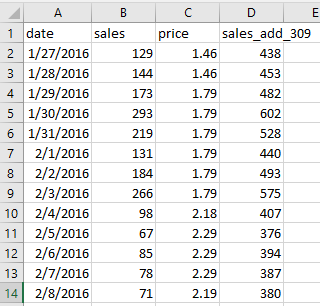}
	}%
	\caption{Adding a constant to training and validation data. We expect the training loss and validation loss to be very similar. The forecast for the validation data is expected to increase approximately by the same constant.}
	\label{mockTrainVaidateData_norm}
\end{figure}

Thus, by adding a constant, the forecast will increase by the same value of the constant. This MR is visualised in Figure \ref{mockTrainVaidateData_norm}.

Similarly, the training data can be subtracted by a constant, or multiplied by a constant (the proof remains similar). For subtraction, the training loss will remain the same (subject to $DefaultVariation$) and the forecast will reduce by the constant. In the case of multiplication, the training loss will remain the same (subject to $DefaultVariation$) and the forecast will be multiplied by the constant.

\subsection{MR-2: corollary to MR-1: linear scaling of only the validation data should significantly alter the forecasts}
Although, at cursory, this test may look obvious, the intention of this test is to ensure that the computation of the $min$ \& $max$ for the normalisation is not done on the validation data (i.e. the $min$ \& $max$ values should be computed from the training data and those values should be used during validation). 

This test is specified as follows. Let the training complete with the original training data. Now, add a large constant (e.g. few multiples of the max value in the training data) to the validation data. The forecast of the first validation data point should alter significantly after removing the constant added. If, on the other hand, the forecast is exactly the same as before, it indicates the normalisation is using values of $min$ \& $max$ for normalisation from the validation data which is incorrect. A similar test can be done by multiplying the validation data by a constant.

\subsection{MR-3: Test that no training data is missed during the generation of sequences}
Generating the set of sequences from the time series data can be tricky and there is a possibility that some data is being missed (particularly around the corner cases). 

\paragraph{Case 1}Truncate the training data such that its length is exactly TIME\_STEPS + NUMBER\_OF\_DAYS\_TO\_FORECAST (i.e. there will be just one sequence to train on). Ensure that the model is able to train. 

\paragraph{Case 2}Truncate the training data such that its length is less than TIME\_STEPS + NUMBER\_OF\_DAYS\_TO\_FORECAST. Ensure that the model does not train and an appropriate error is given.

\paragraph{Case 3}Truncate the training data such that the number of sequences that are generated is less than that of BATCH\_SIZE (a hyper-parameter). This can be done by making the length of training data equal to BATCH\_SIZE + TIME\_STEPS - 1.

\paragraph{Case 4}Truncate the training data such that the number of sequences generated is it exactly equal to BATCH\_SIZE. This can be done by making the length of training data equal to BATCH\_SIZE + TIME\_STEPS.

\subsection{MR-4: Test that the right amount of validation data is used for the generation of sequences}
Similar to the case of creating the sequences for the training data, here we check that the correct amount of validation data is seen to forecast. 

\paragraph{Case 1}Truncate the validation data such that its length is exactly equal to TIME\_STEPS + NUMBER\_OF\_DAYS\_TO\_FORECAST. Ensure that the model is able to give a prediction. 

\paragraph{Case 2}Truncate the validation data to less than TIME\_STEPS + NUMBER\_OF\_DAYS\_TO\_FORECAST. Ensure the model does not give any forecast.

\paragraph{Case 3}Truncate the validation data such that the number of sequences generated is larger than TIME\_STEPS but less than the BATCH\_SIZE. This can be done by making the length of validation data length equal to BATCH\_SIZE + TIME\_STEPS - 1.
 
\paragraph{Case 4}Truncate the validation data such that the number of sequences generated is exactly equal to the hyper-parameter of BATCH\_SIZE. This can be done by making the length of validation data equal to BATCH\_SIZE + TIME\_STEPS.

\subsection{MR-5: Check if time steps are being considered in order}
If there is a possibility that the training data or the validation data may appear out of order, ensure the application is reading the time in ascending order. 

\paragraph{Training} Shuffle the rows of the training data, including time. Ensure post training, the validation loss \& forecasts are approximately same (i.e. subject to default variation).


\paragraph{Validation} Similar to the training data, here we shuffle only the validation data. The results should be exactly the same as before (note this case is not impacted by default variation). 

Both these cases check to see if the application orders the data by time steps (assuming there is a case when the time series data may not be naturally ordered in time).

\subsection{MR-6: Introduce training data with a range of 0}
Since the application normalises the training data where the data is divided by range (see equation \ref{norm}), we introduce an artificial training data which has $0$ range. The application should handle such a case appropriately.

\subsection{MR-7: Introduce validation data with a range of 0}
For the original training data, make an artificial validation data which has 0 range. Ensure the application works normally (i.e. since the range should be computed from the training data, this test should not impact the normal operation of the application).

\subsection{MR-8: Validate the value of TIME\_STEPS used}
The hyper-parameter of TIME\_STEPS denotes the amount of past data that is seen. We choose this value in such a way so as to capture cyclic dependencies in the time series data. However, it isn't clear what value of the TIME\_STEPS should be used. Too small a value, will prevent the network from learning cyclic dependencies. Too large a value will reduce the amount of training data that is used.

We have developed a method to choose the time steps. The idea is to convert the time series data into the frequency domain using Fourier transforms. We then remove two frequency bands which contain the least amount of information. From this truncated data in the frequency domain, we re-construct the time series data and measure the amount of data loss when compared to original data (the data loss is computed as a normalised L2 loss). We then plot this loss against the number of time steps that can be considered. 

\begin{figure}[H]
\centering
	\includegraphics[width=3.5in]{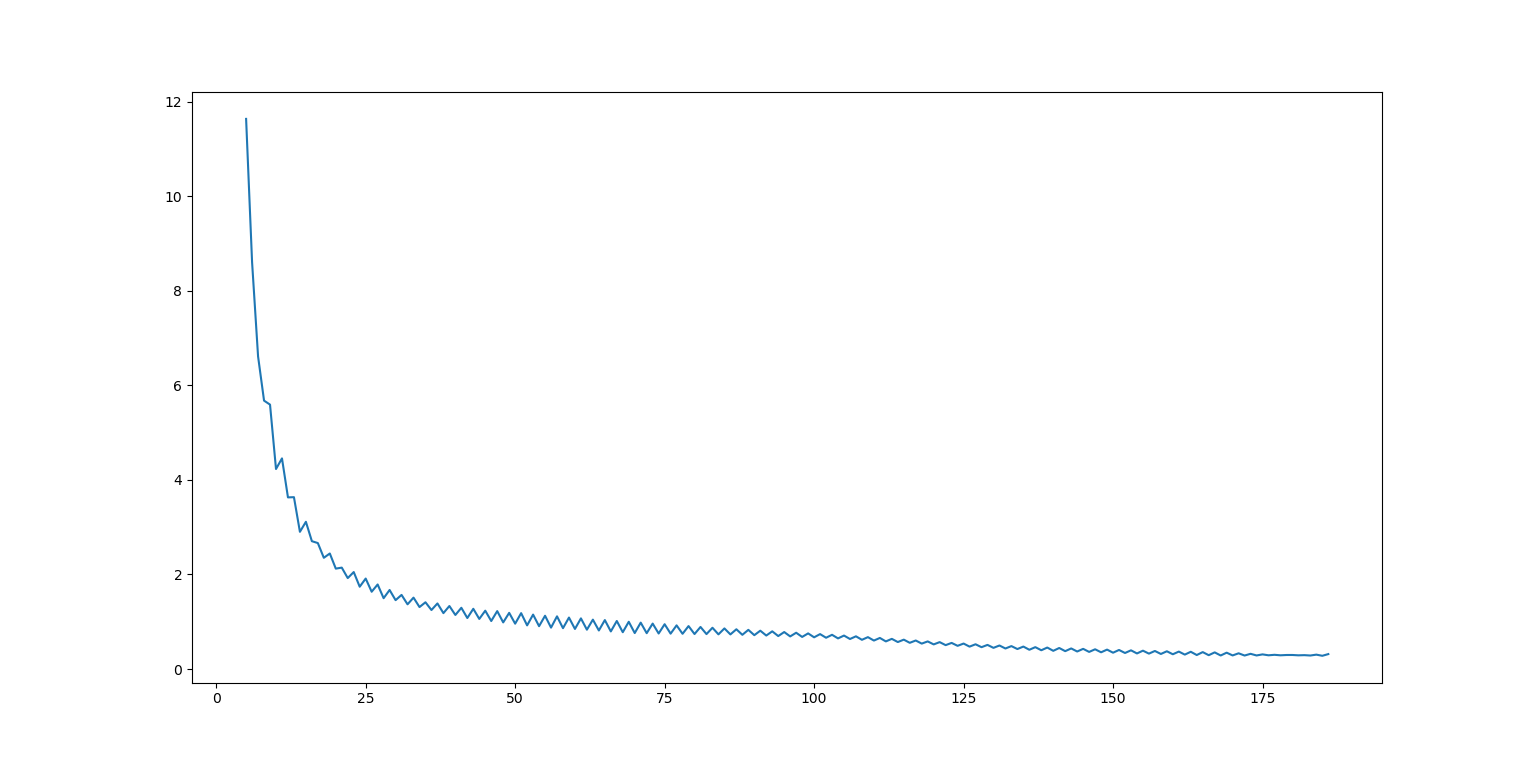}
	\caption{A plot of the number of TIME\_STEPS versus the loss when the frequencies with low information in the data are removed. We choose a TIME\_STEPS around the region where the curve exhibits drastic changes in slope (around 25). }
	\label{recon}
\end{figure}

A sample result of this analysis is shown in Figure \ref{recon}. We choose the TIME\_STEPS around the region where the plot drastically changes its slope towards 0. The algorithm is shown as Algorithm \ref{timeSteps} and the code is released online \footnote{\url{https://github.com/ahujaofficial/Fourier-on-Sequences}}. 

\begin{algorithm}
	\caption{Algorithm to choose the TIME\_STEPS needed.}
	\label{timeSteps}
	\KwIn{Set of Input Validation Data \(\boldsymbol X\)}
	\KwOut{Information loss versus different values of TIME\_STEPS}
	Assign \(Loss=0\) \\
	\For {\(timeStep = 5\) to \(length(X)\)} 
	{
		\ForEach {sequence, \(s\), in \(\boldsymbol X\) of length \(timeStep\)}
		{
			\(f \)= convert \(s\) into frequency domain using Fourier Transform \\
			\(f_{shift}\) = shift \(f\) to center maximum information in the center of the array \\
			\(f_{drop}\) = Assign \(f_{shift}[0] = 0 \) and \(f_{shift}[length(f_{shift)}]=0 \) \tcp{this makes the two frequency bands with least information equal to 0}
			\(f_{invshift}\) = Inverse the shift operation on \(f_{drop}\) \\
			\(s_{reconv}\) = convert \(f_{invshift}\) to time domain using Inverse Fourier Transform \\
			\(l = ||s - s_{reconv}||_2^2 \) \tcp{compute squared L2 loss}
			\(Loss[timeStep] = Loss[timeStep] + \frac{l}{(number\ of\ sequences\ s\ in\ X) * (timeStep)} \)
		}
	}
	Plot \(Loss\) versus the values of \(timeStep\)
\end{algorithm}

\subsection{MR-9: Validate the Robustness of the trained model}
In this test we aim to check whether it is possible to generate adversarial inputs on a model. The problem is defined as follows. Let $X_s$ be a sample input (of length TIME\_STEPS) which gives a forecast of $y_s$. Now we wish to find a new input $X_p$ such that $X_p \approx X_s$ and yet its forecast $y_p = 2*y_s$. If such an input, $X_p$ can be found, it implies a small change in the input can cause a huge change in the output which typically indicates a problem in the training process.

To generate the perturbed input $X_p$, we build an optimisation based method similar in principle to the Carlini \& Wagner method \cite{carlini2017towards}. We provide the algorithm below and the implementation is released online\footnote{ Code here: \url{https://github.com/anuragbms/Sales-forecasting-with-RNNs/blob/master/generateAdversarials.ipynb}}.

\begin{algorithm}
	\caption{Algorithm to generate Adversarial inputs for the LSTM based forecaster application}
	\label{advLSTM}
	\KwIn{Set of Input Validation Data \(\boldsymbol X_s\). Trained Model \(h()\). Number of optimisation steps \(G\)}
	\KwOut{Set of Adversarial Inputs \(\boldsymbol X_p \) such that \(h(X_p)\) = 2 * \(h(X_s)\) and \(X_p \approx X_s\)}
	Define auxiliary variable \(X_{aux} \) of the same shape of \(X_s\)\\
	\(X_p = X_{aux}^2 \) \tcp{To make sure \(X_p\) is always positive}
	\(y_s = h(X_s) \)\\
	\(y_p = h(X_p) \)\\
	\For{each \(X_s \in \boldsymbol X_s\)}
	{
		\(X_{aux} = X_s \)\\
		Define Input Sequence Loss as: \(L_{i} = ||X_s - X_p ||_2^2 \) \tcp{\(||\ \ ||_2^2\) is the squared L2 loss}
		Define Forecast Loss as: \(L_{f} = (y_s - y_p)^2 \) \\
		Define Total loss as: \(L = L_{i} + L_{f} \) \\
		\While{numOfSteps \(<\) G }
		{
			Minimize \( L \) by varying \(X_{aux}\) \tcp{can use ADAM optimiser}
			\(numOfSteps++\)\\
		}
		Add \(X_p\) to \(\boldsymbol X_p \) \\
	}
	Output \(\boldsymbol X_p \)\\
\end{algorithm}

If a large number of adversarial examples are found, it indicates the trained model is not robust.

\section{Results} \label{results}

To measure the efficacy of the MRs and the Metamorphic Testing approach, we have made 2 sets of tests. In the first, we applied the MRs on the in-use OP application and captured the issues discovered. In the second test, we introduced artificial bugs into the reference implementation through Mutation Testing. We then applied the MRs and measured how many bugs were caught. We did not perform Mutation Testing on the OP application since the code was not available with us. Further, Mutation Testing on a reference implementation of the co-relation co-efficient was not done, since the implementation is quite straight forward (bordering on trivial).

\subsection{Results from the OP application}
The OP application uses the co-relation co-efficient (Module 1) and the LSTM based forecaster (Module 2). Each of the MRs were executed on the application by the testing team. The results showed that 8 issues, not known earlier, were identified.

\renewcommand{\arraystretch}{1.5}
\begin{table}[H]
	\caption{Results from the testing of OP application.}
	\label{resultsOP}
	\begin{tabu}{|X[c]|X[c]|X[c]|}
		\hline
		Module & MRs that failed & Number of issues identified \\
		\hline
		1 (co-relation co-efficient) & MR-4, MR-9, MR-10 & 4 \\
		\hline
		2 (LSTM) & MR-5, MR-8, MR-9 & 4 \\
		\hline
	\end{tabu}
\end{table}

\subsection{Results from Mutation Testing}
Hypothetical bugs, in the reference implementation of the LSTM based forecaster, were generated using the concept of Mutation Testing \cite{jia2011analysis}. Mutation Testing systematically changes the original source file by modifying a line of code. For example, if the a line of source code had an operator $<$, Mutation Testing would create a new source file by changing the operator to $>$. Such a modified source file is called as a mutant and numerous mutants for a single source file can be generated. Mutation testing is based on the principle that the mutants generated represent actual errors programmers make \cite{jia2011analysis}. We used the Mutation Testing tool `MutPy' \cite{mutpy} to generate the mutants. 

In total, 403 mutants were created for the reference implementation spanning two source code files. We removed those mutants that result in an exception, or those that change print statements or the hyper-parameters. This resulted in 44 mutants to be valid implementation bugs\footnote{The mutants \& their analysis is here: \url{https://github.com/anuragbms/Sales-forecasting-with-RNNs/blob/master/MetamorphicTests/relevant_mutants.xlsx}}. The MRs were run on each of the mutants and the results are shown in Table \ref{mutResults}. Overall, 29 out of 44 mutants were caught (or 65.9\%).

\begin{table}[H]
	\caption{Results from the testing of the LSTM reference implementation.}
	\label{mutResults}
	\begin{tabu}{|X[c]|X[c]|X[c]|X[c]|}
		\hline
		Soure file & Total Number of valid mutants & MRs that failed & Number of mutants caught \\
		\hline
		Training & 30 & MR-1, MR-3 & 18 \\
		\hline
		Validation & 14 & MR-1, MR-2, MR-3 & 11 \\ 
		\hline
	\end{tabu}
\end{table}

Further details on the results which tabulates the specific mutants and MRs is available online\footnote{\url{https://github.com/anuragbms/Sales-forecasting-with-RNNs/blob/master/MetamorphicTests/results.xlsx}}.

\section{Related Work} \label{related}
Machine Learning (ML) based applications have been traditionally built using the process of `training' and `validation'. Training includes collecting a large amount of data which is used for the algorithm to learn its internal parameters. Validation uses a smaller amount of data but distinct from the set used for training. The difference in the accuracy of the application on the training and validation data-sets is used to judge the performance and the problems in the application. In the current state of the practice, this process of `training' and `validation' is thought to subsume the traditional role of software testing.

However, there are various deficiencies in this standard process of `training' and `validation'. It is sometimes seen that subtle implementation mistakes do not produce obvious signals during training and validation and may go undetected \cite{Dwarakanath} \cite{bugsBen} \cite{selsam2017developing}. It is also often the case that validation data does not represent the true data that would occur in production \cite{liang2017enhancing}.  It is further speculated that, due to some common properties of the training and validation data,  the ML application ends up learning statistical regularities in the data and not the underlying semantic concepts \cite{Jo2017}. The training data may have problems, where certain scenarios are underrepresented or the training algorithm may have issues where certain biases in data may be amplified \cite{Bolukbasi2016}. Finally, the ML algorithm may learn to produce the correct output, but for the wrong reasons \cite{Ribeiro2016}.

All such issues may manifest themselves in such a way that a ML application works well during development, but fails spectacularly during actual use.

Thus, there has been a growing interest in the effective testing of ML based applications. Some of the recent work includes, measuring the invariance of image classifiers to rotations and translations \cite{engstrom2017rotation}, changes in image characteristics such as contrast \cite{geirhos2017comparing} \cite{tian2017deeptest} \cite{pei2017deepxplore} and introducing spurious objects onto an image \cite{rosenfeld2018elephant}. There has been investigation into generation of adversarial inputs for an ML application where the inputs are specifically crafted such that they cause the application into giving a wrong output \cite{szegedy2013intriguing} \cite{papernot2016practical}. Efforts have been made to detect \& mitigate instances of bias in training data \cite{Zhao2017} and an ML algorithm \cite{Bolukbasi2016}. Finally, interpreting the decisions of an ML algorithm has been studied as well \cite{Ribeiro2016} \cite{adebayo2016iterative}.

In this paper, we continue \cite{Dwarakanath} to explore the testing of an ML application with a focus on identifying implementation bugs. We approach the problem through the application of Metamorphic Testing. Some of the existing work in Metamorphic Testing of ML \& statistical applications include the testing of Naive-Bayes classifier \cite{xie2009application}\cite{xie2011testing}, Support Vector Machine with a linear kernel \cite{murphy2010empirical} and k-nearest neighbor \cite{xie2009application} \cite{xie2011testing}. In this paper, we work with a statistical algorithm (Co-relation co-efficient) and a Deep Learning based LSTM network (both of which have not been studied earlier). Further, we also provide results of the approach on an in-use application and the efficacy in catching implementation bugs. 

\section{Conclusion} \label{conclusion}
In this paper, we have presented the Metamorphic Testing of an in-use application that uses statistical analysis and a deep learning based forecasting algorithm. In total, 19 Metamorphic Relations have been created and proofs, where applicable, have been presented. The efficacy of the approach was tested in two ways. In the first, the relations were executed on the application and 8 issues were identified. These issues were spotted even though the application had gone through the standard Machine Learning process of `train' \& `validate'. In the second set of tests, hypothetical bugs were introduced into a reference implementation of the application through the concept of Mutation Testing. The Metamorphic Relations were executed and the number of bugs caught were recorded. Of 44 bugs introduced, the Metamorphic Testing approach caught 29 (or 65.9\%). Overall, the results and the application of the approach in a practical setting showed the ability of Metamorphic Testing to catch issues in statistical and machine learning based applications.


\bibliographystyle{IEEEtran}
\bibliography{ref}

\end{document}